**A Study of Large Language Models for Patient Information Extraction: Model Architecture, Fine-Tuning Strategy, and Multi-task Instruction Tuning**

| | |
|---|---|
| Authors: | Cheng Peng, PhD[1, *]<br>Xinyu Dong, PhD[3, *]<br>Mengxian Lyu, MS[1]<br>Daniel Paredes, MS[1]<br>Yaoyun Zhang, PhD[3, †]<br>Yonghui Wu, PhD[1,2, †] |
| Affiliation of the authors: | [1]Department of Health Outcomes and Biomedical Informatics, College of Medicine, University of Florida, Gainesville, Florida, USA |
| | [2] Preston A. Wells, Jr. Center for Brain Tumor Therapy, Lillian S. Wells Department of Neurosurgery, University of Florida, Gainesville, Florida, USA |
| | [3]Selfii |
| | [*] Equal contribution |
| | [†] Co-corresponding author |
| Corresponding author: | Yaoyun Zhang, PhD<br>Email: yaoyun.zhang@selfii.com |
| | Yonghui Wu, PhD<br>1889 Museum Road, 7th Floor<br>Gainesville, FL, USA, 32611<br>Phone: 352-294-8436<br>Email: yonghui.wu@ufl.edu |

# ABSTRACT

**Background**

Natural language processing (NLP) is a key technology to extract important patient information from clinical narratives to support healthcare applications. The rapid development of large language models (LLMs) has revolutionized many NLP tasks in the clinical domain, yet their optimal use in patient information extraction tasks requires further exploration. This study examines LLMs' effectiveness in patient information extraction, focusing on LLM architectures, fine-tuning strategies, and multi-task instruction tuning techniques for developing robust and generalizable patient information extraction systems.

**Methods**

This study aims to explore key concepts of using LLMs for clinical concept and relation extraction tasks, including: (1) encoder-only or decoder-only LLMs, (2) prompt-based parameter-efficient fine-tuning (PEFT) algorithms, and (3) multi-task instruction tuning on few-shot learning performance. We benchmarked a suite of LLMs, including encoder-based LLMs (BERT, GatorTron) and decoder-based LLMs (GatorTronGPT, Llama 3.1, GatorTronLlama), across five datasets. We compared traditional full-size fine-tuning and prompt-based PEFT. We explored a multi-task instruction tuning framework that combines both tasks across four datasets to evaluate the zero-shot and few-shot learning performance using the leave-one-dataset-out strategy.

**Results**

For single-task clinical CE, the two decoder-based LLMs (Llama 3.1 and GatorTronLlama) achieved the best performance, with average F1 scores of 0.8964 and 0.8981, respectively, across the five datasets, outperforming other LLMs with average F1 improvement of 0.7~3.3%. Encoder-based LLMs with prompt-based learning outperformed those implemented using classification. For clinical RE task, the prompt-based PEFT strategy demonstrated remarkable performance, with an F1 improvement up to 15.9% over traditional fine-tuning on all datasets. All three decoder-based LLMs outperformed encoder-based LLMs, increasing average F1 score by 1.8 to 6.6%, with GatorTronLlama achieved the best performance with an average F1 score of 0.8978. Multi-task instruction tuning showed critical improvements, boosting zero-shot and few-shot F1 scores by 1.1~37.8% compared to those without multi-task fine-tuning. Notably, generative LLMs with multitask instruction tuning using only 20% of the full dataset achieved similar performance comparable to the full-size fine-tuning, with a very small gap less than 0.005 in F1 scores.

**Conclusions**

Our findings support generative LLMs with PEFT as a cost-effective solution for patient information extraction. In addition, we show that multi-task instruction tuning significantly improves the zero-shot and few-shot performance, contributing to better generalizability. This study provides practical guidelines to develop LLM-based scalable, adaptable, and high-performing patient information extraction systems.

# INTRODUCTION

Clinical notes are a valuable component of electronic health records (EHRs), capturing critical clinical care information such as laboratory tests, diagnoses, treatments, and outcomes [1,2]. Information extraction (IE) is the key technology to extract critical patient information to support downstream healthcare applications, such as decision support and clinical trial matching[3–5]. IE comprises two fundamental subtasks, including clinical concept extraction (CCE), which identifies clinical concepts such as diseases, treatments, and symptoms[6], and clinical relation extraction (CRE), which identifies relationships between clinical concepts, such as drug and corresponding adverse-event[7,8]. IE has been extensively explored in previous studies and open challenges, such as the i2b2/n2c2 shared tasks between 2006 and 2012 [9–17]. Many rule-based, machine learning-based, and hybrid methods were developed. Critical challenges of clinical IE identified from previous studies, including the complexity of clinical language, domain-specific vocabulary, as well as high annotation costs. More robust and scalable NLP methodologies are needed[18].

Recent breakthrough in transformer-based large language models (LLMs) have revolutionized many clinical NLP tasks. Early stage models, including rule-based [19]) and traditional machine learning models [20] have limitations in generalizability, recent advancements in LLMs provide promising solutions. The original breakthrough of LLMs came from the encoder-based transformer models, such as BERT[21], which were later customized for biomedical and clinical applications through pre-training using biomedical and clinical corpora, contributing to domain-specific transformer models such as BioBERT[22], ClinicalBERT[23], and GatorTron[24]. While the encoder-based transformer models improved clinical IE through bidirectional, embedding-based text representation, the extraction task was still approached using a classification based extraction, which still suffers the lack of generalizability inherited from the traditional machine learning models [25,26]. After 2020, decoder-based transformer models with massive numbers of parameters became the mainstream, known as generative LLMs, such as ChatGPT, LLAMA [27–29], and GatorTronGPT [30]. Generative LLMs adopted prompt-based learning algorithms instead of classification, where human instructions were integrated to the input as additional information, i.e., prompts, to instruct generative LLMs generating correspondence answers following a text-to-text generation procedure. This offers a more flexible and efficient way to instruct machines to identify required information based on human instructions, thus to formulate multiple NLP tasks in a unified text-to-text generation framework. Another very unique advantage of generative LLMs is the few-shot and zero-shot learning capabilities[31,32], enabling them to achieve human-level language processing with a few even no labeled data and greatly reduced the annotation cost. Fine-tuning is a critical technology to adopt LLMs for clinical IE [8]. Traditional fine-tuning of encoder-based LLMs requires the training of dataset-specific classification layers that requires substantial labeled datasets [33] and the trained classification layer only can be used for one task-specific type of IE. As LLMs typically have massive amounts of parameters typically over billions, it is very expensive to update all parameters during fine-tuning. Parameter-efficient fine-tuning (PEFT), such as P-tuning[34,35] and Low-Rank Adaptation (LoRA)[36], were proposed to reduce the training cost by updating only a small proportion of parameters. Instruction tuning using multi-task data has also demonstrated better few-shot learning capabilities [32,37], compared with fine-tuning using single task.

Both encoder-based and decoder-based LLMs have been applied for clinical IE. However, most recent studies of generative LLMs mainly focused on free text question answers (QA). Several critical gaps exists for clinical IE using LLMs. First, there is a lack of comprehensive comparison between encoder-based LLMs and decoder-based generative LLMs across diverse clinical IE tasks; the strengths and weaknesses of using both LLMs for IE are still not clear. Second, the practical trade-offs between traditional full-size fine-tuning and PEFT among LLM architectures have not been systematically evaluated, leaving practitioners without clear guidance on the selection of fine-tuning strategies. Third, the transfer learning ability of the LLMs to generalize to new datasets and annotation schemas with minimal data—a crucial capability for real-world deployment—remains a significant challenge. The potential of multi-task instruction tuning to specifically enhance the transfer learning and generalizability of LLMs has not been explored for clinical IE. To the best of our knowledge, this is the first study to compare both encoder-based and decoder-based LLMs for patient information extraction focusing on model architectures, fine-tuning strategies, generalizability, and multi-task instruction tuning using extensive clinical data benchmark datasets. Specifically, this study presents a comprehensive empirical study evaluating both encoder-based and decoder-based LLMs for patient information extraction. Our contributions are: (1) provide a benchmark comparison of mainstream encoder-based LLMs (BERT, GatorTron) and decoder-based LLMs (GatorTronGPT, Llama 3.1, GatorTronLlama) for patient information extraction. (2) systematically evaluate the performance and efficiency of traditional fine-tuning versus prompt-based PEFT strategies. (3) examine multi-task instruction tuning in generative LLMs using a leave-one-dataset-out approach to assess the improvement of zero-shot and few-shot transfer learning to unseen datasets. (4) provide an empirically guidance for selecting LLM architectures and fine-tuning strategies to develope scalable, adaptable, and high-performing patient information extraction systems.

## METHODS

This study evaluates two encoder-based LLMs and three decoder-based LLMs for patient information extraction from clinical narratives. We systematically explored the effect of model architecture, fine-tuning strategies, and the generalizability through multi-task instruction tuning. Figure 1 shows an overview of the study design.

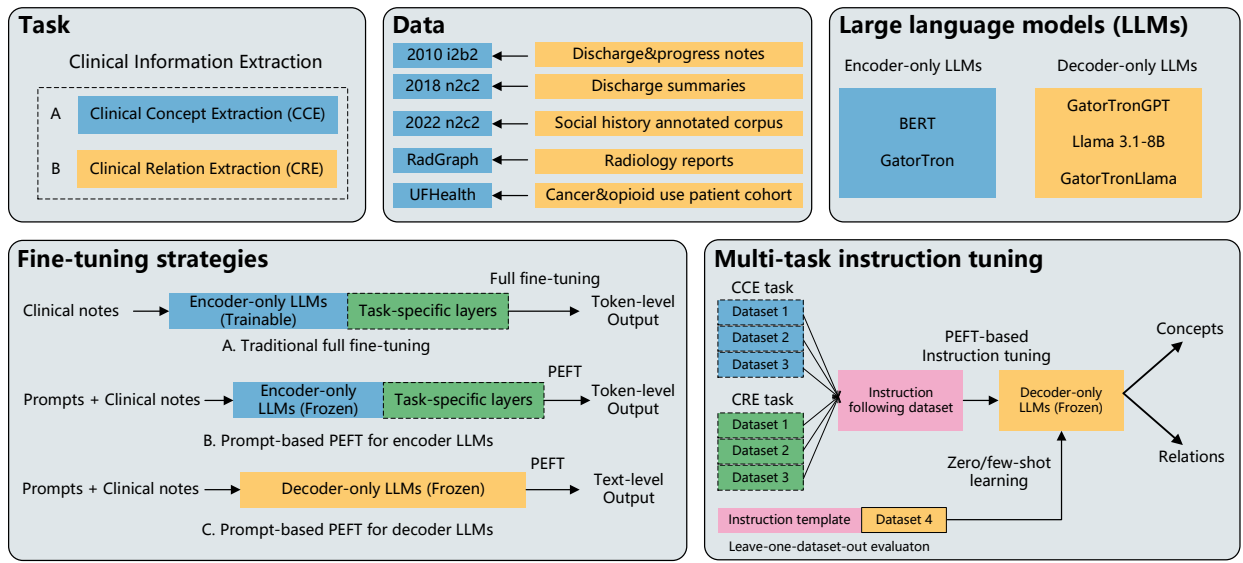

Figure 1 An overview of the study design. PEFT: parameter efficient fine-tuning.

### Clinical NLP tasks

This study focuses on two fundamental NLP tasks for patient information extraction, including clinical CE and clinical RE.

*Clinical CE*, also known as named entity recognition (NER), is the task of identifying and categorizing spans of text that refer to predefined medical concepts, such as diseases, treatments, and medications.

*Clinical RE* focuses on identifying and classifying semantic relationships (e.g., treatment for disease, drug causes adverse event.) between the entities extracted in clinical CE.

**Datasets**

This study utilized five widely used clinical benchmark datasets, including the 2010 i2b2[9], 2018 n2c2[15], 2022 n2c2[17], RadGraph[38], and an internal dataset from UF Health[39]. These corpora span diverse clinical domains, from general clinical notes to specialized areas like radiology and social determinants of health (SDoH), and originate from different institutions with diverse documentation styles and annotation schemas.

*2010 i2b2 dataset*

The 2010 i2b2/VA challenge corpus contains de-identified discharge summaries and progress reports from multiple healthcare institutions. It was annotated for three core medical concept categories, including *Problem*, *Test*, and *Treatment*.

*2018 n2c2 dataset*

This dataset was developed in Track 2 of 2018 n2c2 shared tasks focused on the extraction of medications, and adverse drug events (ADEs) from clinical narratives. It was annotated for 9 categories of clinical concepts (drug, drug attributes, ADEs) and 8 categories of relations among drugs, drug-associated attributes, and ADEs.

*2022 n2c2 dataset*

This dataset was developed in Track 2 of 2022 n2c2 challenge, which focused on the extraction of Social Determinants of Health (SDoH) in social history sections from clinical notes. The 2022 n2c2 dataset consists of 5 categories of SDoH concepts (e.g., employment status, living status, and substance use) and 9 categories of SDoH-associated attribute concepts (e.g., strength, frequency and dosage), and 28 types of relations among SDoH concepts and SDoH-associated attributes (e.g., frequency of substance use).

*Radgraph dataset*

This dataset includes entities and relations in full-text chest X-ray radiology reports from two large-scale datasets MIMIC-CXR and CheXpert. It is annotated with 4 entity types defined by the anatomical location and clinical observation status, and 3 types of relations between entities.

*UF Health dataset*

This dataset developed from our previous study. It collected two disease cohorts from the University of Florida (UF) Health Integrated Data Repository (IDR), including a cancer cohort and an opioid use cohort, which was identified 19 categories of SDoH concepts and 26 categories of relations between SDoH concepts from the clinical notes.

**Table 1** summarizes the five datasets, including their sources, domain, sizes, concept and relation types. By leveraging these diverse datasets, this study aims to provide a comprehensive evaluation of LLMs for patient information extraction, addressing the challenges of generalizability and robustness across different clinical domains, documentation styles, and annotation schemas.

Table 1 Overview of five clinical datasets

| Dataset | Domain | Notes | Concept types/count | Relation types/count | Source |
|---|---|---|---|---|---|
| 2010 i2b2 | Discharge summaries & progress notes | 871 | 3/45,009 | - | 2010 i2b2/VA challenge |
| 2018 n2c2 | Discharge summaries | 505 | 9/83,869 | 8/59,810 | 2018 n2c2 shared task, medication-ADE track |
| 2022 n2c2 | Social History Annotated Corpus (SHAC) | 1877 | 14/21,114 | 28/7,307 | 2022 n2c2 shared task |
| RadGraph | Radiology reports | 600 | 4/17,345 | 3/12,898 | PhysioNet |
| UFHealth | Cancer & Opioid use patient cohorts | 829 | 19/17,535 | 26/6,374 | UF Health IDR |

**LLM architectures**

This study investigates two mainstream LLM architectures, including encoder-based LLMs and decoder-based LLMs.

*Encoder-based LLMs*

Encoder-based LLMs are trained using the encoder component of the transformer architecture, such as BERT. Encoder-based LLMs process the input text bidirectionally, attending to all tokens simultaneously to learn contextual representations. These models are typically pre-trained on a self-supervised objective like Masked Language Modeling (MLM), where the model learns to predict randomly masked tokens within a sentence. For clinical CE, researchers typically add a task-specific classification head (e.g., a linear layer) on top of the final hidden-state. In this study, we evaluated two widely used encoder-only models, including BERT and GatorTron.

- BERT: The first successful bidirectional transformer model that set up the foundation to train for pre-training of transformer models for many NLP tasks.

- GatorTron: An encoder-based clinical LLM adopted the BERT architecture that has been pre-trained on a large-scale corpus of over 90 billion words of clinical and biomedical text, including de-identified notes from the UFHealth system, making it a powerful clinical foundational LLM.

*Decoder-based LLMs*

Decoder-based LLMs, also known as generative LLMs, are typically trained using the encoder component of the transformer architecture, also known as generative LLMs, such as the GPT. Decoder-based LLMs utilize a unidirectional (or autoregressive) transformer architecture. They are pre-trained on a Causal Language Modeling (CLM) objective, where the model learns to predict the next token in a sequence given all preceding tokens. A very unique characteristic of the decoder-based LLMs is that they can solve multiple NLP tasks using a unified text-to-text learning framework, which is not feasible for encoder-based LLMs. Instead of adding task-specific layers (i.e., encoder-based LLMs), generative LLMs are guided by human instructions to generate the desired answers. This study evaluated three decoder-only models, including GatorTronGPT, Llama 3.1 and GatorTronLLAMA.

- GatorTronGPT: A generative clinical LLM developed in our previous work, which is pre-trained using 277 billion words of text comprising 82 billion words of clinical text and 195 billion words of diverse general English text.
- Llama 3.1: A general-domain foundation model from Meta AI, was pretrained on a massive, diverse corpus, offering scalability and robust performance across generative tasks.
- GatorTronLlama: A specialized model that leverages the Llama 3.1-8B architecture and adapts it with deep clinical domain knowledge through continued pre-training using over 100 billion words of clinical text collected from UF Health.

**LLM fine-tuning strategies**

Fine-tuning is a supervised learning algorithm to adopt pretrained LLMs for specific NLP tasks such as patient information extraction, using a small set of annotated corpus. This study examines two fine-tuning strategies, including traditional full-size fine-tuning – where all parameters were optimized, and PEFT – only a very small proportion of parameters were updated. The choice of strategy was aligned with the model's architecture, allowing for a robust comparison between established and modern adaptation techniques.

*Traditional Full-size Fine-Tuning*

This approach was widely used to fine-tune encoder-based models (e.g., BERT and GatorTron). For clinical CE, a dataset-specific classification layer, which is a linear classification layer applied to each token's final hidden-state, was added to predict its concept label (e.g., B-Problem, I-Problem, O). For clinical RE, we first generate candidate concept pairs and use a binary classification head to determine whether the candidate pair has a relation or not. During training, the entire model, including all transformer layers and the new classification layers, are updated. This method is computationally expensive for LLMs with billions of parameters and require large datasets to prevent overfitting.

*Parameter-Efficient Fine-Tuning (PEFT)*

We adopted a prompt-based Machine Reading Comprehension (MRC) framework developed in our previous work to fine-tune encoder-based LLMs, where clinical concepts and relations were extracted by using human instructions as prompts. For example, we used the following question, "find the drug events including names, brand names and collective names" as a prompt for MRC models to identify the drug mentions, which were then used to generate relation-related questions as new prompts to find other concepts (e.g., the "Strength" of drugs). To identify the answer, we used two binary classifiers to predict the span (start and end indexes) of the entities. For the decoder-only generative LLMs (GatorTronGPT, Llama 3.1, and GatorTronLlama), we reformulated the two extraction tasks into a unified, text-to-text format, allowing the model to process them using a natural language instruction. For example, we added a prompt "Extract all medical problems from the following clinical note" for the 2010 i2b2 CCE task and a question "Does the relationship between drug event 'Metformin' and frequency 'daily' exists?" for the 2018 n2c2 CRE task. Instead of classifying tokens, the model is prompted to generate the structured information directly as a text string. Which provides more flexibility, allowing both CE and RE can be handled by the same LLM. We utilized LoRA, a parameter efficient fine-tuning framework, to reduce computational cost of fully fine-tuning LLMs with multi-billion parameters. LoRA injects small, trainable low-rank matrices into the attention layers of the Transformer architecture. During fine-tuning, only the low-rank matrices are updated.

**Multi-task instruction tuning**

Multi-task instruction tuning is a technology to enhance the generalizability of LLMs by training them using a mixed dataset containing multiple tasks, with the assumption that the knowledge learned from one task could help other tasks. In this study, we applied multi-task instruction tuning for decoder-bases LLMs (GatorTronGPT, Llama 3.1, and GatorTronLlama). Multitask instruction tuning can only be applied to LLMs that can model different tasks using a unified architecture, such as generative LLMs. For each task, multiple human designed prompt templates, which vary in phrasing and structure, are employed to specify the desired output format. These prompts, examples of which are shown in supplementary Table S1, are integrated with our LoRA-based PEFT framework, which updates a small subset of model parameters while adapting to both tasks.

To evaluate the effectiveness of multi-task instruction tuning, we adopt a leave-one-dataset-out approach for few-shot and zero-shot learning experiments. In this setup, the model is trained on three of the four datasets (2018 n2c2, 2022 n2c2, RadGraph, and UFHealth datasets), with the remaining dataset held out for evaluation of the few-shot and zero-shot performance.

**Experimental design**

We conducted single-task benchmarking experiments to compare the selected LLMs for single-task settings and conducted multitask instruction experiments to compare the performance for few-shot performance.

*Single-task Benchmarking*

We finetuned each selected LLM using a single dataset setting to test the performance. Seven LLMs with different sizes were compared, including BERT-large (340 million), GatorTron-base (345 million), GatorTron-large (9 billion), GatorTronGPT-base (5 billion), GatorTronGPT-large (20 billion), Llama 3.1-8B (8 billion), and GatorTronLlama (8 billion). Each model is trained and evaluated on each of the five datasets independently. This experiment allows for a controlled comparison of: (1) encoder-only vs. decoder-only architectures, (2) traditional vs. PEFT fine-tuning, and (3) the impact of model sizes on task performance.

*Multi-Task Instruction Tuning*

Three generative LLMs, including GatoTronGPT-base, Llama 3.1-8B and GatorTronLlama were explored. We used a leave-one-dataset-out protocol using the 2018 n2c2, 2022 n2c2, RadGraph, and UF Health datasets. This setup involves four folds, using the four datasets. In each fold, one dataset is held out for evaluation under zero-shot and few-shot settings. Zero-shot is assessed by applying the multi-task trained model directly to the held-out dataset without further fine-tuning, while few-shot performance is evaluated using the model fine-tuning on small subsets (5, 10, 20, 50 samples). The performance of these multi-task models is compared with single-dataset fine-tuned models.

*Evaluation metrics*

The micro-averaged F1-score based on strict match was used for evaluation of CE, which requires an exact match for both the entity type and the start and end boundaries of the entity span. Similarly, the micro-averaged F1-score was used to evaluate RE for classifying the relation type between pairs of concepts.

*Implementation Details*

All experiments using encoder-only models (BERT, GatorTron) were conducted using the Hugging Face transformers library and the PyTorch Lightning framework. For full-size fine-tuning, we used learning rate of 2e-5 with the AdamW optimizer, and training batch size of 8. Experiments for all decoder-only generative models (GatorTronGPT, Llama 3.1, GatorTronLlama) were performed using the NVIDIA NeMo framework. All prompt-based PEFT was conducted within NeMo. For these models, a learning rate of 1e-4 was used with the AdamW optimizer. The specific LoRA configuration included a rank (adapter_dim) of 256 and a dropout of 0.2 applied to the adapter layers. All experiments were conducted using four Nvidia A100-80G GPUs.

RESULTS

Table 2 compares 9 LLMs with different architectures and fine-tuning strategies for clinical CE using single task finetuning. Three LLMs, including GatorTron-large-MRC, Llama 3.1-8B and GatorTronLlama achieved top performance with average F1 score over 0.89 across five datasets. GatorTronLlama achieved the highest average F1-score of 0.8981, narrowly outperforming the general-domain Llama 3.1-8B (0.8964) and the prompt-based encoder GatorTron-large-MRC (0.8946), while the smaller GatorTron-base-MRC models had the lowest average F1 score of 0.8653, with the lowest score observed from the 2022 n2c2 dataset (0.8186) compared with other models. Among fine-tuning strategies for encoder-based models, smaller GatorTron-base-MRC with LoRA-based PEFT (0.8653) slightly underperformed its full fine-tuned counterpart (0.8711) by 0.58%, while larger GatorTron-large-MRC achieved average F1 score of 0.8946, significantly outperforming the smaller GatorTron-base-MRC by 2.9%, and its fully fine-tuned counterpart (0.8807) by 1.39%. Among prompt-based models, decoder-only LLMs GatorTronGPT-base and GatorTronGPT-large achieved F1 scores of 0.8882 and 0.8899, respectively, slightly underperformed encoder-only GatorTron-large-MRC, while other two decoder-only LLMs, Llama 3.1-8B (0.8964) and GatorTronLlama-8B (0.8981), outperformed encoder-only models, surpassing GatorTron-base-MRC by 3.11% and 3.28%. Notably, all models consistently scored lowest on the 2022 n2c2 dataset, while GatorTronLlama showed its best performance on the RadGraph and UF Health datasets, with F1-scores of 0.9179 and 0.9199, respectively.

Table 2. Performance comparison of LLMs with different architectures and fine-tuning strategies for clinical concept extraction.

| Model | Architecture | Params | Fine-tuning strategy | Dataset (micro average F1 score) | | | | | |
|---|---|---|---|---|---|---|---|---|---|
| | | | | 2010 i2b2 | 2018 n2c2 | 2022 n2c2 | RadGraph | UF Health | Average |
| BERT-large | Encoder-only | 340M | | 0.8694 | 0.8807 | 0.8318 | 0.8852 | 0.9048 | 0.8744 |
| GatorTron-base | | 345M | Full fine-tuning | 0.8715 | 0.8817 | 0.8341 | 0.8869 | 0.9114 | 0.8771 |
| GatorTron-large | | 9B | | 0.8782 | 0.8825 | 0.8388 | 0.8915 | 0.9125 | 0.8807 |
| GatorTron-base-MRC | | 345M | Prompt-based parameter efficient fine-tuning | 0.8705 | 0.8612 | 0.8186 | 0.8818 | 0.8945 | 0.8653 |
| GatorTron-large-MRC | | 9B | | 0.8953 | 0.8993 | 0.8548 | 0.9081 | 0.9153 | 0.8946 |
| GatorTronGPT-base | Decoder-only | 5B | | 0.8784 | 0.9021 | 0.8462 | 0.9021 | 0.9122 | 0.8882 |
| GatorTronGPT-large | | 20B | | 0.8806 | 0.9024 | 0.8511 | 0.9025 | 0.9129 | 0.8899 |
| Llama 3.1-8B | | 8B | | **0.8903** | 0.9035 | 0.8561 | 0.9136 | 0.9187 | 0.8964 |
| GatorTronLlama | | 8B | | 0.8862 | **0.9054** | **0.8612** | **0.9179** | **0.9199** | **0.8981** |

**Table 3** compares 9 LLMs with different architectures and fine-tuning strategies for clinical RE. Llama 3.1-8B and GatorTronLlama-8B with prompt-based PEFT are the best models achieved comparable average F1-scores of 0.8942 and 0.8978, respectively, with exceptional performance on 2018 n2c2 dataset (0.9612 and 0.9674) and UF Health dataset (0.9085 and 0.9124). BERT-large and GatorTron-large with traditional fine-tuning contributed the lowest average F1-scores around 0.83, particularly struggling on RadGraph dataset (0.6985 and 0.6925). For encoder-only models, two GatorTron models with prompt-based fine-tuning (GatorTron-base-MRC) achieved average scores of 0.8506 and 0.8661, outperforming their classification-based fine-tuned counterparts (0.8322 and 0.8375) by 1.84% and 2.84%, respectively. Decoder-only models with prompt-based PEFT significantly outperformed encoder-only models, with GatorTronGPT-large (0.8862), Llama 3.1-8B (0.8942), and GatorTronLlama-8B (0.8978) achieving 3.36%~4.72% higher F1-scores than the best encoder-only model GatorTron-large-MRC (0.8661). The performance on the 2018 n2c2 dataset was remarkably high across all models, over 0.95, while RadGraph consistently presented the lowest scores of 0.6985~0.8514.

**Table 3**. Performance comparison of LLMs with different architectures and fine-tuning strategies for clinical relation extraction

| Model | Architecture | Params | Fine-tuning strategy | Dataset (micro average F1 score) | | | | |
|---|---|---|---|---|---|---|---|---|
| | | | | 2018 n2c2 | 2022 n2c2 | RadGraph | UFHealth | Average |
| BERT-large | Encoder-only | 340M | Full fine-tuning | 0.9561 | 0.7807 | 0.6985 | 0.8952 | 0.8326 |
| GatorTron-base | | 345M | | 0.9575 | 0.7798 | 0.6925 | 0.8988 | 0.8322 |
| GatorTron-large | | 9B | | 0.9580 | 0.7838 | 0.7058 | 0.9025 | 0.8375 |
| GatorTron-base-MRC | | 345M | Prompt-based parameter efficient fine-tuning | 0.9358 | 0.8452 | 0.7358 | 0.8855 | 0.8506 |
| GatorTron-large-MRC | | 9B | | 0.9605 | 0.8489 | 0.7552 | 0.8997 | 0.8661 |
| GatorTronGPT-base | Decoder-only | 5B | | 0.9635 | 0.8520 | 0.8182 | 0.9024 | 0.8840 |
| GatorTronGPT-large | | 20B | | 0.9657 | 0.8529 | 0.8205 | 0.9056 | 0.8862 |
| Llama 3.1-8B | | 8B | | 0.9612 | **0.8615** | 0.8454 | 0.9085 | 0.8942 |
| GatorTronLlama | | 8B | | **0.9674** | 0.8601 | **0.8514** | **0.9124** | **0.8978** |

The zero-shot and few-shot learning experiments evaluate the generalization capabilities of three decoder-only large language models (LLMs)—GatorTronGPT, Llama 3.1-8B, and GatorTronLlama-8B—using a leave-one-dataset-out approach across four datasets: 2018 n2c2, 2022 n2c2, RadGraph, and UFHealth datasets. Figure 2 compares zero-shot and few-shot learning performance of three decoder-based LLMs, including GatorTronGPT, Llama 3.1-8B, and GatorTronLlama-8B, using four datasets, including 2018 n2c2, 2022 n2c2, RadGraph, and UFHealth datasets. The average F1-scores across the four datasets was used for evaluation. For zero-shot performance, multi-task instruction tuned GatorTronLlama-8B achieved the highest average F1-scores of 0.3596 for CE and 0.3154 for RE, significantly outperforming its counterpart without instruction tuning (0.0155 for CE, 0.0189 for RE) by 34.41% and 29.65%, respectively. All multi-task tuned models, including GatorTronGPT-base, Llama 3.1 and GatorTronLlama, achieved significant improvement with averages F1 scores of 0.2658~0.3955, while all baseline models without instruction tuning performed poorly across the two tasks, with average F1 scores near random chance or close to zero. In few-shot settings (5, 10, 20 samples), multi-task tuned GatorTronLlama-8B achieved the highest F1-scores of 0.7350 (CE) and 0.6815 (RE) at 5 shots,

0.7570 (CE) and 0.7115 (RE) at 10 shots, and 0.7811 (CE) and 0.7685 (RE) at 20 shots, outperforming baselines without instruction tuning by 9.7%~12.6% for CE and 5.9%~8.6% for RE. The three multi-task models consistently outperform their single-task baselines on both CE and RE tasks, achieving performance improvement of 0.8~9.7% for CE, and 1.8~6.0% for RE. While the performance gap between multi-task tuned and baseline models narrows as more examples are provided, the multi-task tuned models consistently take the lead. The Llama 3.1-8B and GatorTronLlama models, when enhanced with multi-task tuning, show comparable and robust few-shot learning performance. Notably, performance on CE is consistently higher than RE across all models. For example, in the 20-shot setting, the multi-task tuned GatorTronLlama achieves an F1-score of 0.7811 for CE, compared to 0.7685 for RE.

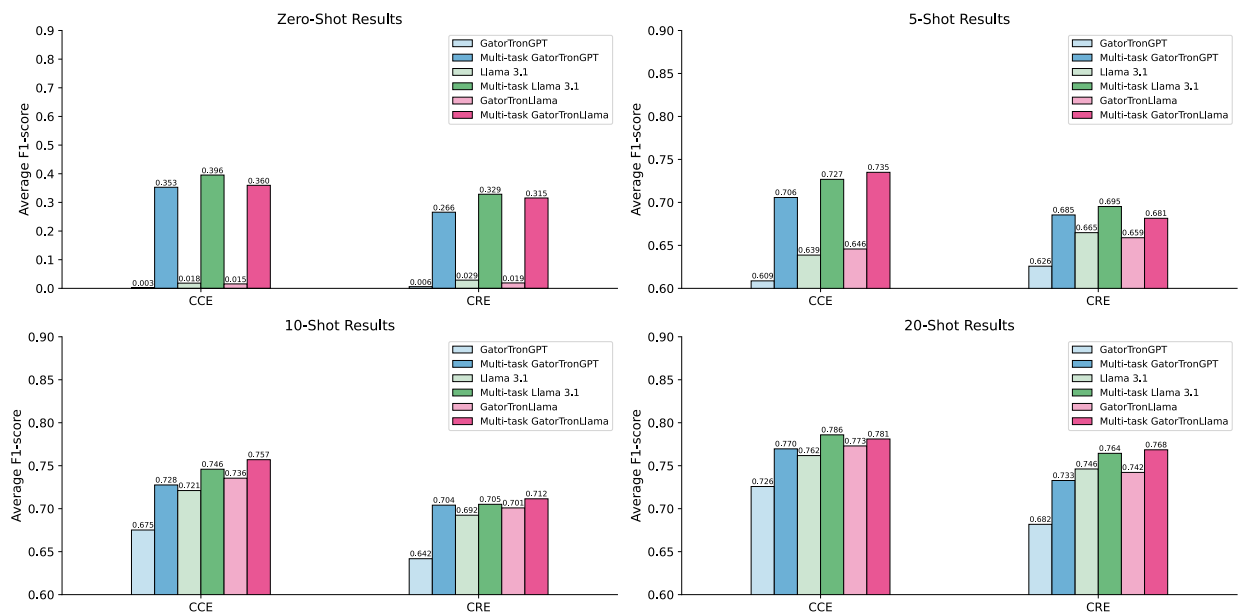

Figure 2. Zero-shot and few-shot performance between multi-task instruction tuned LLMs and baseline models without instruction tuning across CCE and CRE tasks.

**DISCUSSION**

This study provided a comprehensive evaluation of LLMs for patient information extraction, including CE and RE tasks. We systematically compared two LLM architectures, fine-tuning strategies, and the impact of multi-task instruction tuning on generalizability. This study provides key insights into the development of robust and adaptive clinical NLP systems for patient information extraction.

While encoder-based LLMs fine-tuned using advanced MRC framework remain competitive for CE tasks, decoder-based generative LLMs tuned with prompt-based PEFT matched or exceeded the best encoder-only models with full fine-tuning or prompt-based PEFT. Our results support the advantage of decoder-based generative LLMs for patient information extraction.

This study provides guidance on the selection of fine-tuning strategies and their associated computational costs. As shown in Table 4, LoRA-based PEFT presents a better trade-off between performance and efficiency. Compared to traditional full-size fine-tuning, which requires updating all parameters, LoRA trains less than 1% of the total parameters, reducing significantly the training time and GPU memory requirements, making the adaptation of multi-billion parameter models computationally affordable. For example, fully fine-tuning the 9-billion parameter GatorTron-large model required approximately 48 GPU hours, whereas fine-tuning the comparable GatorTronLlama-8B with LoRA took only 8 GPU hours. In addition, the achieved efficiency does not affect the inference speed; because the LoRA adapter weights can be merged into the base model's weights, there is few additional latencies during inference. Notably, the GatorTron-base with full fine-tuning performed better than GatorTron-base-MRC, indicating that for smaller models, updating all parameters can still be the best solution if sufficient in-domain data are available. However, for generative LLMs with much large sizes, prompt-based PEFT is a better choice, with higher performance and efficiency.

**Table 4.** Comparison of computational efficiency for different LLMs.

| Model | Params | Fine-Tuning Strategy | Trainable Params | Average Training Time (GPU Hours) | Average Inference Time (ms/note) |
|---|---|---|---|---|---|
| GatorTron-base | 345 million | Full fine-tuning | 100% | ~8 | ~15 |
| GatorTron-large | 9 billion | | | ~48 | ~22 |
| GatorTron-base-MRC | 345 million | Prompt-based LoRA | ~0.5% | ~2 | ~16 |
| GatorTron-large-MRC | 9 billion | | | ~6 | ~25 |
| GaotTronLlma | 8 billion | | ~0.8% | ~8 | ~32 |

A significant contribution of this work is that multi-task instruction tuning greatly improves few-shot learning for better generalizability. Our leave-one-dataset-out experiments show that LLMs fine-tuned using single dataset have limited transfer learning ability when applied to a new, unseen domain (zero-shot). Multitask instruction tuning remarkably improved the performance. Our findings support the hypothesis that through multitask instruction tuning, generative LLMs can learn knowledge with better generalizability to other tasks. To further illustrate, Figure 3 shows the performance trajectory as we fine-tune the Llama 3.1 and GatorTronLlama models on an increasing percentage of the held-out training data. The models exhibit a steep learning curve, rapidly approaching the performance of a fully fine-tuned baseline. As shown in Figure 3, by using only around 20% of the available training data, the multi-task models achieve F1-scores comparable to models trained using the full datasets.

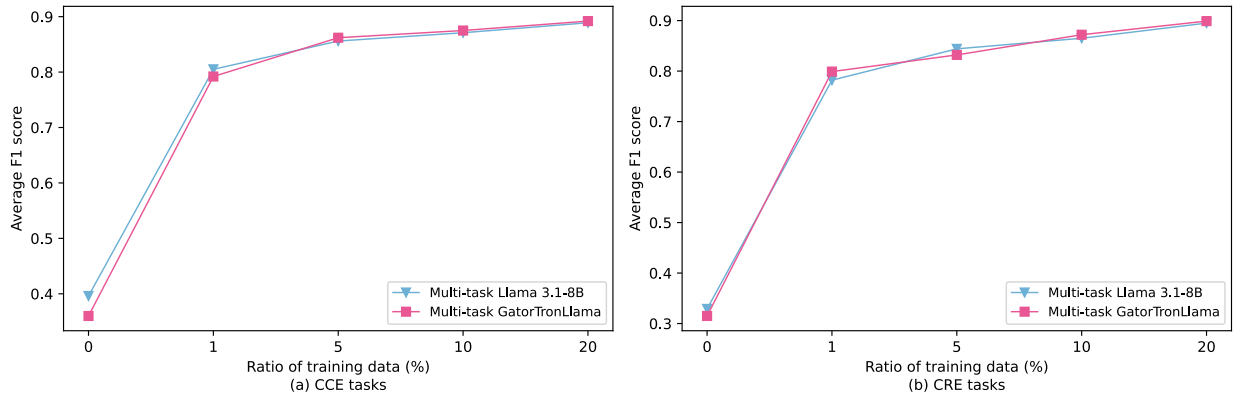

Figure 3 Performance trend with increasing training samples between multi-task instruction tuned Llama 3.1-8B and GatorTronLlama across CCE and CRE tasks

We conducted an error analysis on a small subset of the test data across all datasets. For CE task, the most common errors across all models were boundary errors (e.g., extracting "severe headache" instead of "headache"), especially in the 2022 n2c2 datasets with nested entities related to SDoH (e.g., "Currently unemployed" overlapped with "unemployed") and misclassification errors (e.g., labeling a drug as a Problem). Encoder-only models, due to their token-level classification nature, were unable to identify overlapping or discontinuous entities, a limitation the generative models largely overcome. Although decoders were less prone to these errors, they may produce duplicate spans (e.g., the same concepts occurred multiple times). In RE, errors included misclassification of relations, such as mistaking a "Suggestive Of" relationship for an "Located At", and false positives, where binary classification instructions were employed. Additionally, in multi-task tuning evaluation, errors related to domain shift were most prominent across all datasets because the testing data were from a held-out dataset that the model had not been exposed to during fine-tuning.

This study has limitations. We explored commonly used LLMs and fine-tuning strategies, new algorithms recently proposed need further investigation. The quality of prompts heavily impact the prompt-based solutions. We identified the best prompts for each task based on prompt engineering to the best limit of our enumeration capacity; future studies need to explore advanced prompt engineering techniques.

## CONCLUSION

This study presents a comprehensive study of using encoder-based and decoder-based LLMs for patient information extraction focusing on model architectures, fine-tuning strategies, and multi-task instruction tuning of generative LLMs. Our findings provide a practical guideline in selecting LLM architectures and fine-tuning strategies to facilitate developing LLM-based solutions for patient information extraction from clinical narratives.

## ACKNOWLEDGEMENTS

This study was partially supported by grants from Patient-Centered Outcomes Research Institute® (PCORI®) Award (ME-2018C3-14754, ME-2023C3-35934), the PARADIGM program awarded by the Advanced Research Projects Agency for Health (ARPA-H), National Institute on Aging, NIA R56AG069880, National Institute of Allergy and Infectious Diseases, NIAID R01AI172875, National Heart, Lung, and Blood Institute, R01HL169277, National Institute on Drug Abuse, NIDA R01DA050676, R01DA057886, National Cancer Institute, NCI R37CA272473, National Library of Medicine, NLM R01LM011934, and the UF Clinical and Translational Science Institute. We also acknowledge the contributions of Xinyu Dong and Yaoyun Zhang, employees of Selfii Company. The content is solely the responsibility of the authors and does not necessarily represent the official views of the funding institutions. We gratefully acknowledge the support of NVIDIA Corporation and the NVIDIA AI Technology Center (NVAITC) UF program.

## COMPETING INTERESTS STATEMENT

Cheng Peng, Xinyu Dong, Yaoyun Zhang, Mengxian Lyu, Daniel Paredes, and Yonghui Wu have no conflicts of interest that are directly relevant to the content of this study.

## CONTRIBUTORSHIP STATEMENT

CP, XD, YZ, and YW were responsible for the overall design, development, and evaluation of this study. CP, XD, YZ, ML and DP performed the experiments. CP and YW did the initial drafts of the manuscript, XD, YZ, ML, and DP also contributed to the writing and editing of this manuscript. All authors reviewed the manuscript critically for scientific content, and all authors gave final approval of the manuscript for publication.

## SUPPLEMENTARY MATERIAL

Attached in a separate document.

## DATA AVAILABILITY

Four of the five datasets used in this study are publicly available. The 2010 i2b2, 2018 n2c2, and 2022 n2c2 challenge datasets can be accessed through the n2c2 website (https://n2c2.dbmi.hms.harvard.edu/data-sets). The RadGraph dataset is available for download from PhysioNet (https://physionet.org/content/radgraph/1.0.0/) .

The UF Health dataset was created internally using clinical notes from the University of Florida Health Integrated Data Repository and is not publicly available due to UF policies.